\documentclass[runningheads]{llncs}
\usepackage{xcolor}
\usepackage[TU]{fontenc}
\pagecolor{white}

\usepackage{cite}
\usepackage{amsmath,amssymb,amsfonts}
\usepackage[ruled]{algorithm2e}
\usepackage{graphicx}
\usepackage{textcomp}
\usepackage{xcolor}
\usepackage{booktabs}
\usepackage[hidelinks]{hyperref}
\usepackage{caption}
\usepackage{subcaption}
\usepackage{academicons}
\usepackage{orcidlink}
\usepackage{float}
\usepackage{listings}
\SetKwInput{KwInput}{Input}   
\SetKwInput{KwOutput}{Output}
\SetKwProg{KwProc}{Procedure}{}{}
\SetKwComment{Comment}{/* }{ */}

\newcommand{\orcid}[1]{\href{https://orcid.org/#1}{\textcolor[HTML]{A6CE39}{\aiOrcid}}}
\begin{document}

\title{Two Models are Better than One: Federated Learning Is Not Private For Google GBoard Next Word Prediction}
\titlerunning{Two Models are Better than One}

\author{Mohamed Suliman\orcidlink{0000-0001-8097-6297}\thanks{Now at IBM Research Europe - Dublin.} \and
Douglas Leith\orcidlink{0000-0003-4056-4014}}
\authorrunning{M. Suliman and D. Leith}
%
\institute{Trinity College Dublin, The University of Dublin, College Green, Dublin 2, D02 PN40, Ireland\\
\email{\{sulimanm,doug.leith\}@tcd.ie}}

\maketitle

\begin{abstract}
In this paper we present new attacks against federated learning when
used to train natural language text models.  We illustrate the
effectiveness of the attacks against the next word prediction model
used in Google's GBoard app, a widely used mobile keyboard app that
has been an early adopter of federated learning for production use.
We demonstrate that the words a user types on their mobile handset,
e.g. when sending text messages, can be recovered with high accuracy
under a wide range of conditions and that counter-measures such a use
of mini-batches and adding local noise are ineffective.  We also show
that the word order (and so the actual sentences typed) can be
reconstructed with high fidelity.  This raises obvious privacy
concerns, particularly since GBoard is in production use.
\end{abstract}

\section{Introduction}
Federated Learning (FL) is a class of distributed algorithms for the
training of machine learning models such as neural networks.  A
primary aim of FL when it was first introduced was to enhance user
privacy, namely by keeping sensitive data stored locally and avoiding
uploading it to a central server~\cite{mcmahan2017communication}.  The
basic idea is that users train a local version of the model with their
own data, and share only the resulting model parameters with a central
coordinating server.  This server then combines the models of all the
participants, transmits the aggregate back to them, and this cycle
(i.e. a single FL `round') repeats until the model is judged to have
converged.

A notable real-world deployment of FL is within Google's Gboard, a
widely used Android keyboard application that comes pre-installed on
many mobile handsets and which has $>5$ Billion
downloads~\cite{gboardplay22}.  Within GBoard, FL is used to train the
Next Word Prediction (NWP) model that provides the suggested next
words that appear above the keyboard while
typing~\cite{hard2018federated}.

In this paper we show that FL is not private for Next Word
Prediction. We present an attack that reconstructs the original
training data, i.e. the text typed by a user, from the FL parameter
updates with a high degree of fidelity.  Both the FedSGD and
FederatedAveraging variants of FL are susceptible to this attack.  In
fairness, Google have been aware of the possibility of information
leakage from FL updates since the earliest days of FL, e.g. see
footnote 1 in~\cite{mcmahan2017communication}.  Our results
demonstrate that not only does information leakage indeed happen for
real-world models deployed in production and in widespread use, but
that the amount of information leaked is enough to allow the local
training data to be fully reconstructed.

We also show that adding Gaussian noise to the transmitted updates,
which has been proposed to ensure local Differential Privacy (DP),
provides little defence unless the noise levels used are so large that
the utility of the model becomes substantially degraded.  That is, DP
is not an effective countermeasure to our attack\footnote{DP aims to
  protect the aggregate training data/model against query-based
  attacks, whereas our attack targets the individual updates.
  Nevertheless, we note that DP is sometimes suggested as a potential
  defence against the type of attack carried out here.}. We also show
that use of mini-batches of up to 256 sentences provides little
protection. Other defences, such as Secure Aggregation (a form of
multi-party computation MPC), Homomorphic Encryption (HE), and Trusted
Execution Environments (TEEs), are currently either impractical or
require the client to trust that the server is
honest\footnote{Google's Secure Aggregation
  approach~\cite{bonawitz2016practical} is a prominent example of an
  approach requiring trust in the server, or more specifically in the
  PKI infrastructure which in practice is operated by the same
  organisation that runs the FL server since it involves
  authentication/verification of clients.  We note also that Secure
  Aggregation is not currently deployed in the GBoard app despite
  being proposed 6 years ago.} in which case use of FL is redundant.

Previous studies of reconstruction attacks against FL have mainly
focused on image reconstruction, rather than text as considered here.
Unfortunately, we find that the methods developed for image
reconstruction, which are based on used gradient descent to minimise
the distance between the observed model data and the model data
corresponding to a synthetic input, do not readily transfer over to
text data.  This is perhaps unsurprising since the inherently discrete
nature of text makes the cost surface highly non-smooth and so
gradient-based optimisation is difficult to apply successfully.  In
this paper we therefore propose new reconstruction approaches that are
specialised to text data.

It is important to note that the transmission of user data to a remote
server is not inherently a breach of privacy. The risk of a privacy
breach is related to the nature of the data being sent, as well as
whether it's owner can be readily identified. For example, sending
device models, version numbers, and locale/region information is not
an immediate concern but it seems clear that the sentences entered by
users, e.g. when typing messages, writing notes and emails, web
browsing and performing searches, may well be private. Indeed, it is
not only the sentences typed which can be sensitive but also the set
of words used (i.e. even without knowing the word ordering) since this
can be used for targeting surveillance via keyword blacklists
\cite{guardiannsasms}.

In addition, most Google telemetry is tagged with with an Android
ID. Via other data collected by Google Play Services the Android ID is
linked to (i) the handset hardware serial number, (ii) the SIM IMEI
(which uniquely identifies the SIM slot) and (iii) the user's Google
account \cite{infocomgaen21,securecom21}.  When creating a Google
account users are encouraged to supply a phone number and for many
people this will be their own phone number.  Use of Google services
such as buying a paid app on the Google Play store or using Google Pay
further links a person's Google account to their credit card/bank
details.  A user's Google account, and so the Android ID, can
therefore commonly be expected to be linked to the person's real
identity.


\section{Preliminaries}
\subsection{Federated Learning Client Update}
Algorithm \ref{alg:fedlearning} gives the procedure followed by FL
participants to generate a model update. The number of local
epochs $E$, mini-batch size $B$, and local client learning rate $\eta$
can be changed depending on the FL application. When $E = 1$ and the
mini-batch size is equal the size of the training dataset, then it is
called FedSGD, and any other configuration corresponds to
FedAveraging, where multiple gradient descent steps occur on the
client.
\begin{algorithm}
  \caption{Federated Learning Client Update}\label{alg:fedlearning}
  \KwInput{$\theta_0$: The parameters of the global model, training loss function $\ell$.}
  \KwOutput{$\theta_1$: The model parameters after training on the client's data.}  
  \KwProc{clientUpdate}{
    $\theta_1 \gets \theta_0$\;
    $\mathcal{B} \gets (\mbox{split dataset into batches of size } B)$\;
    $\mbox{\textbf{for} local epoch \textit{i} from 1 to $E$ \textbf{do}}$\;
    $\ \ \mbox{\textbf{for} batch \textit{b} $\in \mathcal{B}$ \textbf{do}}$\;
    $\ \ \ \ \theta_1 \gets \theta_1 - \eta\nabla \ell(\theta_1;b)$\; 
    \Return $\theta_1$\;
  }
\end{algorithm}

\subsection{Threat Model}
The threat model is that of a honest-but-curious adversary that has 
access to (i) the FL model architecture, (ii) the current global FL model parameters $\theta_0$, and (iii) the 
FL model parameters, $\theta_1$, after updating locally using the training data of an individual user.   The FL server has, for example, access to all of these and so this threat model captures the situation where there is an honest-but-curious FL server.

We do not consider membership attacks against the global model, although knowledge of $\theta_0$ allows such attacks, since they have already received much attention in the literature.   Instead we focus on local reconstruction attacks i.e. attacks that aim to reconstruct the local training data of a user from knowledge of $\theta_0$ and $\theta_1$.   In the GBoard Next Word Prediction task the local training data is the text typed by the user while using apps on their mobile handset e.g. while sending text messages.


\subsection{GBoard NWP Model}
\begin{figure}[htbp]
\centering
\includegraphics[width=0.4\textwidth]{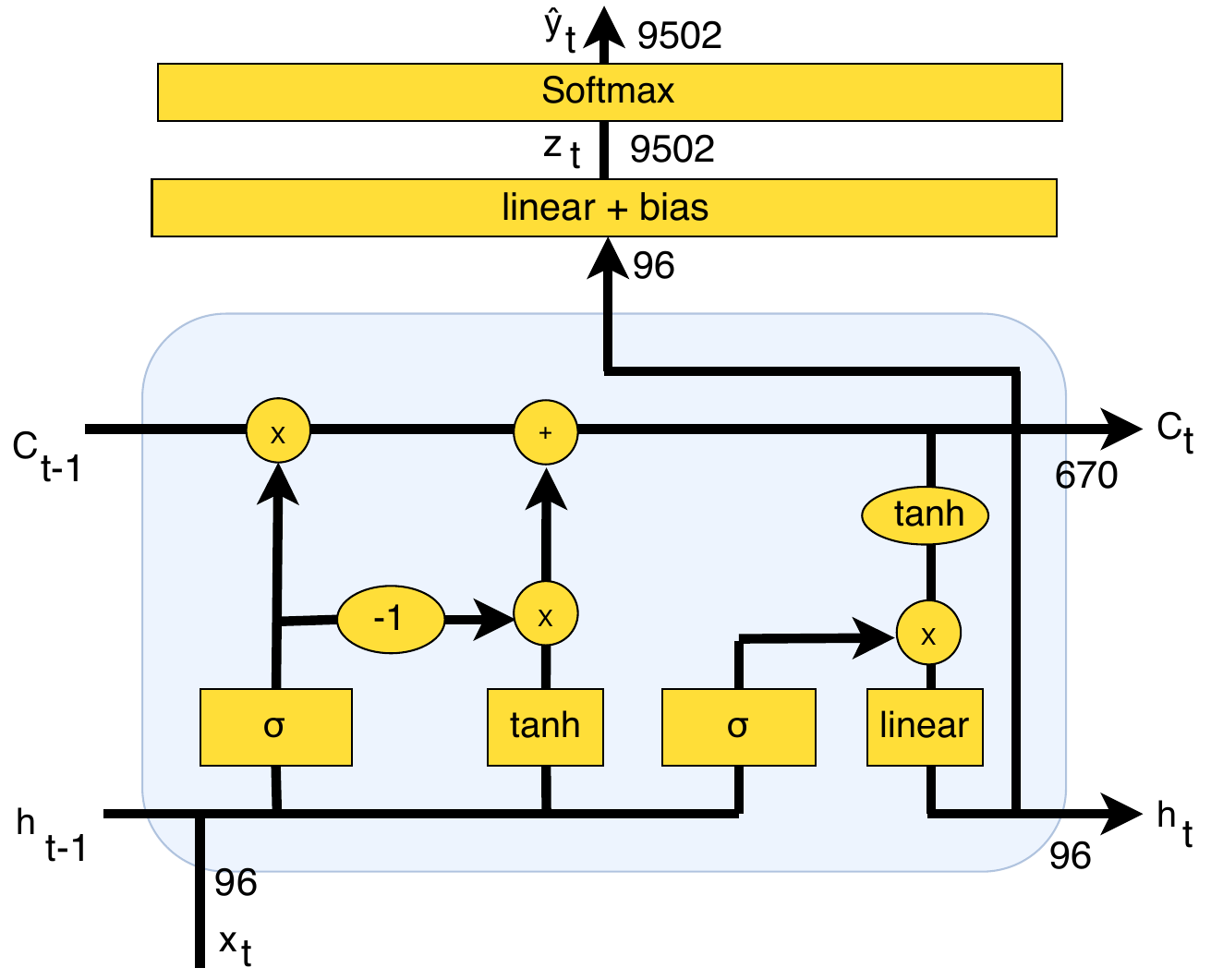}
\caption{\label{fig:gboard_lstm}Schematic of LSTM architecture.  LSTM
  layer takes as input dense vector \(x_t\) representing a typed word
  and outputs a dense vector \(h_t\).  This output is then mapped to
  vector \(z_t\) of size 9502 (the size of the dictionary) with the
  value of each element being the raw logit for the corresponding
  dictionary word.  A softmax layer then normalises the raw \(z_t\)
  values to give a vector \(\hat{y}_t\) of probabilities.}
\end{figure}

Figure \ref{fig:gboard_lstm} shows the LSTM recursive neural net (RNN) architecture used by by Gboard for NWP.   This model was extracted from the app.


The Gboard LSTM RNN is a word level language model, predicting the
probability of the next word given what the user has already typed
into the keyboard. Input words are first mapped to a dictionary entry,
which has a vocabulary of \(V = 9502\) words, with a special
\texttt{<UNK>} entry used for words that are not in the dictionary,
and \texttt{<S>} to indicate the start of a sentence.  The index of
the dictionary entry is then mapped to a dense vector of size \(D=96\)
using a lookup table (the dictionary entry is one-hot encoded and then
multiplied by a \(\mathbb{R}^{D \times V}\) weighting matrix \(W^T\))
and applied as input to an LSTM layer with 670 units i.e. the state
\(C_t\) is a vector of size 670.  The LSTM layer uses a CIFG
architecture without peephole connections, illustrated schematically
in Figure \ref{fig:gboard_lstm}. The LSTM state \(C_t\) is linearly
projected down to an output vector \(h_t\) of size \(D\), which is
mapped to a raw logit vector \(z_t\) of size \(V\) via a weighting
matrix \(W\) and bias \(b\). This extra linear projection is not part
of the orthodox CIFG cell structure that was introduced in
\cite{cifg}, and is included to accommodate the model's tied input and
output embedding matrices \cite{press2017using}. A softmax output
layer finally maps this to an \([0,1]^{V}\) vector \(\hat{y}_t\) of
probabilities, the \(i\)'th element being the estimated probability
that the next word is the \(i\)'th dictionary entry.

\section{Reconstruction Attack}
\subsection{Word Recovery}
\label{sec:wr}

In next word prediction the input to the RNN is echoed in it's output.
That is, the output of the RNN aims to match the sequence of words
typed by the user, albeit with a shift one word ahead.  The sign of
the output loss gradient directly reveals information about the words
typed by the user, which can then be recovered easily by
inspection. This key observation, first made in \cite{zhao2020idlg},
is the basis of our word recovery attack.

After the user has typed \(t\) words, the output of the LSTM model at
timestep \(t\) is the next word prediction vector \(\hat{y}_{t}\),
\begin{align*} \hat{y}_{t,i} = \frac{e^{z_i}}{\sum_{j=1}^Ve^{z_j}},\
i=1,\dots, V
\end{align*} with raw logit vector \(z_t=Wh_t+b\), where \(h_t\) is
the output of the LSTM layer.  The cross-entropy loss function for
text consisting of \(T\) words is \(J_{1:T}(\theta)= \sum_{t=1}^T
J_t(\theta)\) where
\begin{align*} J_t(\theta)= -\log
\frac{e^{z_{i^*}(\theta)}}{\sum_{j=1}^Ve^{z_j(\theta)}},
\end{align*} and \(i^*_t\) is the dictionary index of the \(t\)'th
word entered by the user and \(\theta\) is the vector of neural net
parameters (including the elements of \(W\) and \(b\)).
Differentiating with respect to the output bias parameters \(b\) we
have that,
\begin{align*} \frac{\partial J_{1:T}}{\partial b_{k}} = \sum_{t=1}^T
\sum_{i=1}^V \frac{\partial J_t}{\partial z_{t,i}} \frac{\partial
z_{t,i}}{\partial b_k}
\end{align*} where
\begin{align*} \frac{\partial J_t}{\partial
z_{t,i_t^*}}=\frac{e^{z_{i^*}}}{\sum_{j=1}^Ve^{z_j}}-1<0, \\ \quad
\frac{\partial J_t}{\partial
z_{t,i}}=\frac{e^{z_{i}}}{\sum_{j=1}^Ve^{z_j}}>0,\ i\ne i_t^*
\end{align*} and
\begin{align*} \frac{\partial z_{t,i}}{\partial b_k} = \begin{cases} 1
& k=i\\ 0 & \text{otherwise}
\end{cases}
\end{align*} That is,
\begin{align*} \frac{\partial J_{1:T}}{\partial b_{k}} = \sum_{t=1}^T
\frac{\partial J_t}{\partial z_{t,k}}
\end{align*} It follows that for words \(k\) which do not appear in
the text \(\frac{\partial J_{1:T}}{\partial b_{k}} >0\).  Also,
assuming that the neural net has been trained to have reasonable
performance then \(e^{z_k}\) will tend to be small for words \(k\)
that do not appear next and large for words which do.  Therefore for
words \(i^*\) that appear in the text we expect that \(\frac{\partial
J_{1:T}}{\partial b_{i^*}} <0\).

The above analysis focuses mainly on the bias parameters of the final
fully connected layer, however similar methods can be applied to the
\(W\) parameters. The key aspect here that lends to the ease of this
attack is that the outputs echo the inputs, unlike for example, the
task of object detection in images. In that case, the output is just
the object label.

This observation is intuitive from a loss function minimisation
perspective.  Typically the estimated probability \(\hat{y}_{i^*}\) for
an input word will be less than 1.  Increasing \(\hat{y}_{i^*}\) will
therefore decrease the loss function i.e. the gradient is negative.
Conversely, the estimated probability \(\hat{y}_{i}\) for a word that
does not appear in the input will be small but greater than 0.
Decreasing \(\hat{y}_{i}\) will therefore decrease the loss function
i.e. the gradient is positive.

\emph{Example:} To execute this attack in practice, simply subtract
the final layer parameters of the current global model $\theta_0$ from
those of the resulting model trained on the client's local data,
$\theta_1$, as shown in Algorithm \ref{alg:word-extraction}. The
indices of the negative values reveal the typed words. Suppose the
client's local data consists of just the one sentence ``learning
online is not so private''. We then train model $\theta_0$ on this
sentence for 1 epoch, with a mini-batch size of 1, and SGD learning
rate of 0.001 (FedSGD), and report the the values at the negative
indices in Table \ref{tab:vals}.

\begin{table}
  \centering
  \begin{tabular}{ |c|c|c| }    
    \hline
    word & $i$ & $(\theta_1 - \theta_0)_i $ \\
    \hline
    learning & 7437 & -0.0009951561 \\ 
    online & 4904 & -0.0009941629 \\ 
    is & 209 & -0.000997875 \\
    not & 1808 & -0.0009941144 \\
    so & 26 & -0.0009965639 \\
    private & 6314 & -0.0009951561 \\    
    \hline
  \end{tabular}
  \caption{Values of the final layer parameter difference at the
    indices of the typed words. Produced after training the model on
    the sentence ``learning online is not so private'',
    $E = 1, B = 1, \eta = 0.001$. These are the only indices where
    negative values occur.\label{tab:vals}}
\end{table}

\begin{algorithm}
  \caption{Word Recovery}\label{alg:word-extraction}
  \KwInput{$\theta_0$: The global model's final layer parameters,
    $\theta_1$: The final layer parameters of a model update}
  \KwOutput{User typed tokens $w$}
  \KwProc{recoverWords}{
    $d \gets \theta_1 - \theta_0$\; $w \gets \{ i \ | \ d_i < 0 \}$\;
    \Return $w$\;
  }
\end{algorithm}

\subsection{Reconstructing Sentences}
\label{sec:rs}

The attack described previously retrieves the words typed, but gives
no indication of the order in which they occurred. To reveal this
order, we ask the model\footnote{It is perhaps worth noting that we
  studied a variety of reconstruction attacks, e.g., using Monte Carlo
  Tree Search to perform a smart search over all words sequences, but
  found the attack method described here to be simple, efficient and
  highly effective}.

The basic idea is that after running multiple rounds of gradient
descent on the local training data, the local model is ``tuned'' to
the local data in the sense that when it is presented with the first
words of a sentence from the local training data, the model's next
word prediction will tend to match the training data and so we can
bootstrap reconstruction of the full training data text.

In more detail, the $t$'th input word is represented by a vector
$x_t\in{0,1}^V$, with all elements zero apart from the element
corresponding to the index of the word in the dictionary.  The output
$y_{t+1}\in[0,1]^V$ from the model after seeing input words
$x_0,\dots,x_t$ is a probability distribution over the dictionary.  We
begin by selecting $x_0$ equal to the start of sentence token
\texttt{<S>} and $x_1$ equal to the first word from our set of
reconstructed words, then ask the model to generate
$y_2=Pr(x_{2} | x_0,x_1; \theta_1)$.  We set all elements of $y_2$
that are not in the set of reconstructed words to zero, since we know
that these were not part of the local training data, re-normalise $y_2$
so that its elements sum to one, and then select the most likely next
word as $x_2$.  We now repeat this process for
$y_3=Pr(x_{3} | x_0,x_1,x_2; \theta_1)$, and so on, until a complete
sentence has been generated.  We then take the second word from our
set of reconstructed words as $x_1$ and repeat to generate a second
sentence, and so on.

This method generates as many sentences as there are extracted
words. This results in a lot more sentences than were originally in
the client's training dataset. In order to filter out the unnecessary
sentences, we rank each generated sentence by its change in
perplexity, from the initial global model $\theta_0$ to the new update
$\theta_1$.

The Log-Perplexity of a sequence $x_0,...,x_t$, is defined as
\begin{equation*}
PP_\theta(x_0,...,x_t) =  \sum_{i=1}^{t}(- \log Pr(x_i | x_0,...,x_{i-1}; \theta)),
\end{equation*}
and quantifies how `surprised' the model is by the sequence. Those
sentences that report a high perplexity for $\theta_0$ but a
comparatively lower one for $\theta_1$ reveal themselves as having
been part of the dataset used to train $\theta_1$. Each generated
sentence is scored by their percentage change in perplexity:
\begin{equation*}
  Score(x_0,...,x_t) = \frac{PP_{\theta_0}(x_0,...,x_t) - PP_{\theta_1}(x_0,...,x_t)}{PP_{\theta_0}(x_0,...,x_t)}.
\end{equation*}
By selecting the top-$n$ ranked sentences, we select those most likely
to have been present in the training dataset.

\begin{figure*}[htb]
  \centering
  \begin{subfigure}[b]{0.4\textwidth}
    \centering
    \includegraphics[width=\textwidth]{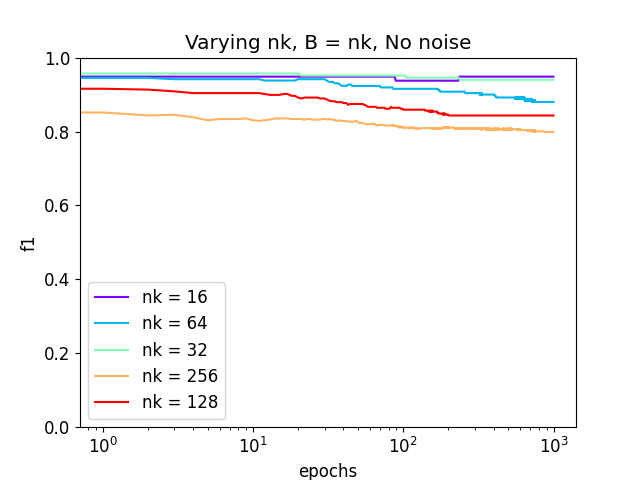}
    \caption{}
   \label{fig:fbepoch}
  \end{subfigure}  
  \begin{subfigure}[b]{0.4\textwidth}
    \centering
    \includegraphics[width=\textwidth]{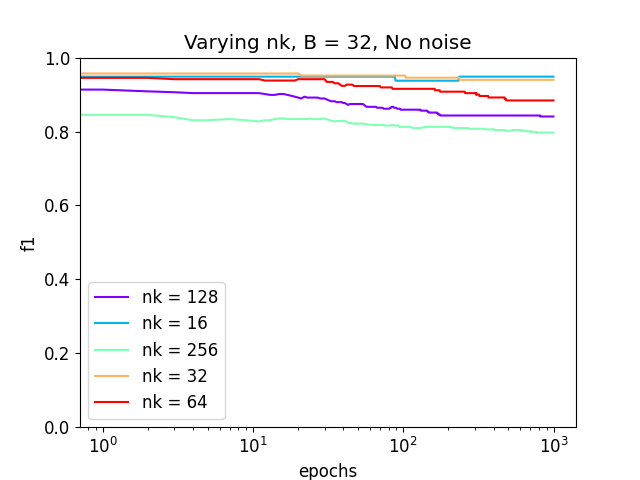}
    \caption{}
    \label{fig:mbepoch}
  \end{subfigure}
  \caption{\small Word recovery performance over time with full batch
    and mini-batch training. The upper bound on the F1 score for the different
    datasets is related to how many words in the training set are also
    in the model dictionary. Figure \ref{fig:fbepoch} shows the F1
    score over time with $B = n_k$ full batch training. At epoch 1
    (FedSGD), the F1 is high and stays constant as you train for
    longer (FedAveraging). Using a mini-batch $B = 32$ (Figure
    \ref{fig:mbepoch}) has no effect on the attack (in the case where
    $n_k = 16, B = 16$).}
  \label{fig:wr_nonoise}
\end{figure*}

\section{Performance Of Attacks Against Vanilla Federated Learning}
\label{sec:results_vanilla}
\subsection{Experimental Setup}
We make use of the LSTM RNN extracted from Gboard as the basis of our
experiments. The value of it's extracted parameters are used as the
initial 'global' model $\theta_0$; the starting point of the updates
we generate. There are several variables that go into producing an
update: the number of sentences in the dataset, $n_k$, the number of
epochs $E$, the batch size $B$, and the local learning rate
$\eta$. Note that when $E = 1$ and $B = n_k$, this corresponds to a
FedSGD update, and that any other configuration corresponds to a
FedAveraging update. Unless explicitly mentioned otherwise, we keep
the client learning rate $\eta = 0.001$ constant for all our
experiments. All sample datasets used consist of 4 word long
sentences, mirroring the average length of sentences that the Gboard
model was trained with \cite{hard2018federated}.

To evaluate the effectiveness of our attack, the sample datasets we
use are taken from a corpus of american english text messages
\cite{oday2013text}, which includes short sentences similar to those
the Gboard LSTM extracted from the mobile application was trained
on. We perform our two attacks on datasets consisting of
$n_k = 16, 32, 64, 128,$ and $256$ sentences. Converting a sentence
into a sequence of training samples and labels $(\textbf{x}, y)$ is done as follows:
\begin{itemize}
\item The start-of-sentence token \texttt{<S>} is prepended to the
  beginning of the sentence, and each word is then converted to it's
  corresponding word embedding.  This gives a sequence
  $x_0,x_1,\dots,x_T$ of tokens where $x_0$ is the \texttt{<S>} token.
\item A sentence of length $T$ becomes $T$ training points

\begin{equation*}
  ((x_0,x_1),y_2), ((x_0,x_1,x_2),y_3), \dots, ((x_0,x_1,\dots,x_{T-1}),y_T)
\end{equation*}
  
  where label $y_t\in [0,1]^V$ is a probability distribution over the
  dictionary entries, with all elements zero apart from the element
  corresponding to the dictionary index of $x_t$

\end{itemize}

Following \cite{hard2018federated} we use categorical cross entropy
loss over the output and target labels. After creating the training
samples and labels from a dataset of $n_k$ sentences, we train model
$\theta_0$ on this training data for a specified number of epochs $E$
with a mini-batch size of $B$, according to Algorithm
\ref{alg:fedlearning} to produce the local update $\theta_1$. We then
subtract the final layer parameters of the two models to recover the
words, and iteratively sample $\theta_1$ according to the methodology
described in Section \ref{sec:rs} to reconstruct the sentences, and
take the top-$n_k$ ranked sentences by their perplexity score.

\subsection{Metrics}
To evaluate performance, we use the F1 score which balances the
precision and recall of word recovery with our attack. We also use a
modified version of the Levenshtein ratio i.e. the normalised
Levenshtein distance \cite{marzal1993computation} (the minimum number
of word level edits needed to make one string match another) to
evaluate our sentence reconstruction attack. Ranging from 0 to 100,
the larger the Levenshtein ratio, the closer the match between our
reconstructed and the ground truth sentence.  


\subsection{Measurements}

\subsubsection{Word Recovery Performance}
Figure \ref{fig:wr_nonoise} shows the measured performance of our word
recovery attack for both the FedSGD and FedAveraging variants of FL
for mini-batch/full batch training and as the dataset size and
training time are varied.  It can be seen that none of these variables
have much of an effect on the F1 score achieved by our attack, which
remains high across a wide range of conditions.  Note that the maximum
value of the F1 score is not one for this data but instead a smaller
value related to how many of the words in the dataset are also present
in the model's vocabulary. Some words, e.g. unique nouns, slang, etc,
do not exist in the model's 9502 word dictionary, and our word
recovery attack can only extract the \texttt{<UNK>} token in their
place, limiting how many words we can actually recover.

\begin{figure*}
  \centering
  \begin{subfigure}[b]{0.325\textwidth}
    \centering
    \includegraphics[width=\textwidth]{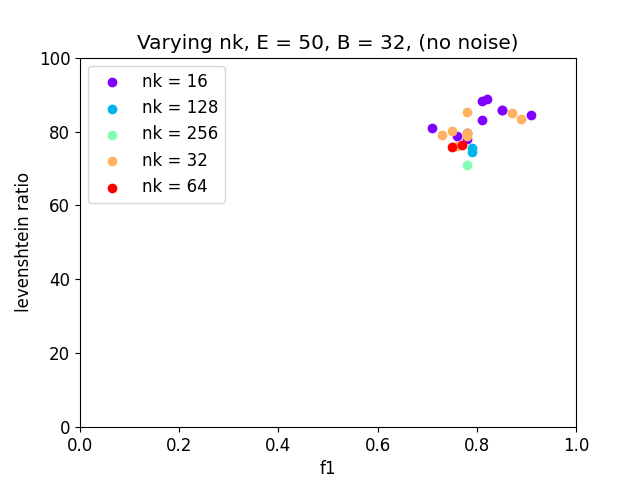}
    \caption{$E = 50$ epochs}
    \label{fig:sre50}
  \end{subfigure}
  \begin{subfigure}[b]{0.325\textwidth}
    \centering
    \includegraphics[width=\textwidth]{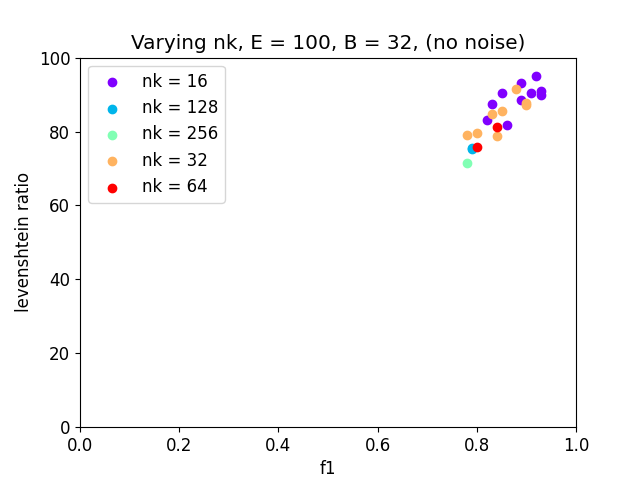}
    \caption{$E = 100$ epochs}
    \label{fig:sre100}
  \end{subfigure}
  \begin{subfigure}[b]{0.325\textwidth}
    \centering
    \includegraphics[width=\textwidth]{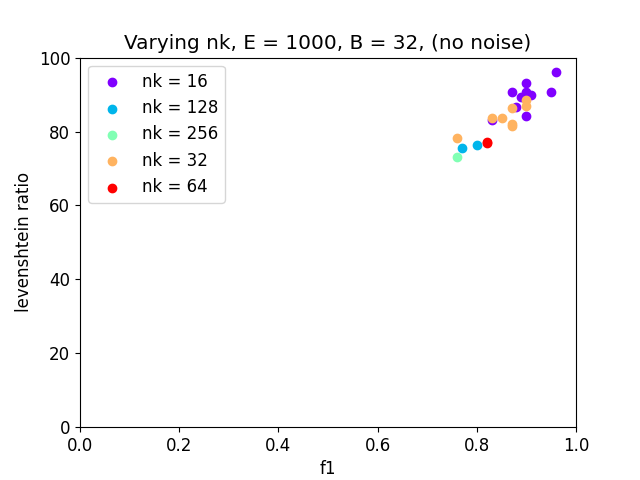}
    \caption{$E = 1000$ epochs}
    \label{fig:sre1000}
  \end{subfigure}  
  \caption{\small Sentence reconstruction performance. Each point
    corresponds to a different dataset colour coded by it's size. The
    y-axis gives the average Levenshtein ratio of the reconstructed
    sentences. The x-axis is the F1 score between the tokens used in
    the reconstructed sentences and the ground truth. The closer a
    point is to the top-right corner, the closer the reconstruction is
    to perfect.}
  \label{fig:fedavg_sr_nonoise}
\end{figure*}

\subsubsection{Sentence Reconstruction Performance}
Figure \ref{fig:fedavg_sr_nonoise} shows the measured performance of
our sentence reconstruction attack. Figures \ref{fig:sre50},
\ref{fig:sre100}, and \ref{fig:sre1000} show that as you train for
more epochs (50, 100, and 1000 respectively) the quality of the
reconstructed sentences improves. This is intuitive as the models
trained for longer are more overfit to the data, and so the iterative
sampling approach is more likely to return the correct next word given
a conditioning prefix.

\begin{figure*}
  \centering
  \begin{subfigure}[b]{0.4\textwidth}
    \centering
    \includegraphics[width=\textwidth]{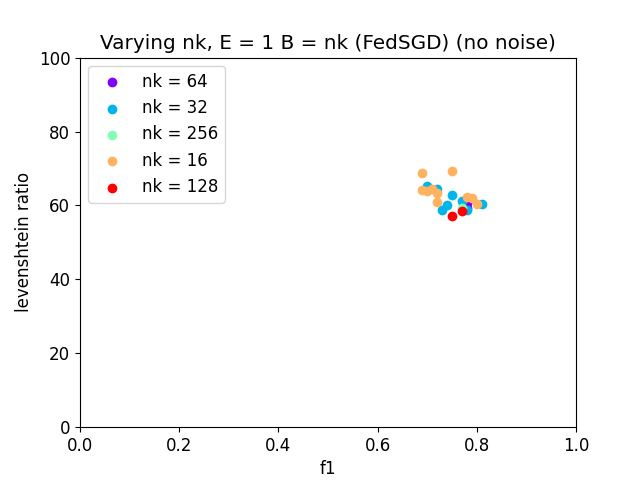}
    \caption{}
    \label{fig:fedsgd_no_scale}
  \end{subfigure}
  \begin{subfigure}[b]{0.4\textwidth}
    \centering
    \includegraphics[width=\textwidth]{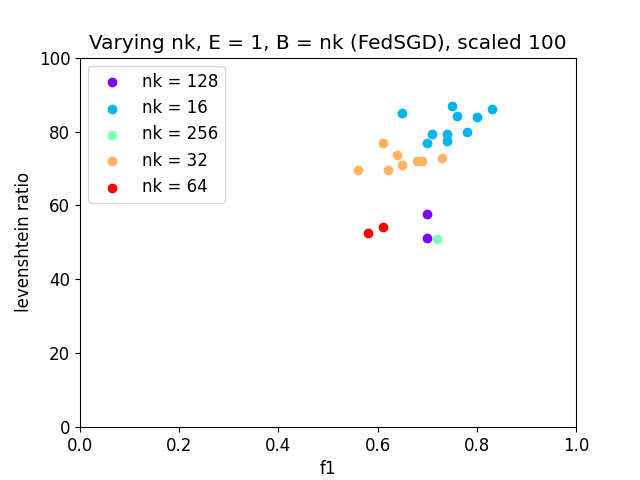}
    \caption{}
    \label{fig:fedsgd_scale}
  \end{subfigure}
  \caption{\small FedSGD sentence reconstruction performance without (a) and with (b) scaling.}
  \label{fig:fedsgd_sr_nonoise}
\end{figure*}

However, longer training times are not necessary to accurately
reconstruct sentences. It can be seen from Figure
\ref{fig:fedsgd_sr_nonoise}(b) that even in the FedSGD setting, where
the number of epochs $E = 1$, we can sometimes still get high quality
sentence reconstructions by modifying the model parameters $\theta_1$
to be $\theta_1 + s(\theta_1 - \theta_0)$, with $s$ being a scaling
factor.  Since $\theta_1 = \theta_0 -\eta \nabla \ell(\theta_0;b)$
with FedSGD,
$$\theta_1 + s(\theta_1 - \theta_0) = \theta_0 -\eta(1+s) \nabla
\ell(\theta_0;b)$$ from which it can be seen that scaling factor $s$
effectively increases the gradient descent step size.


\section{Existing Attacks And Their Defences}
\subsection{Image Data Reconstruction}
Information leakage from the gradients of neural networks used for
object detection in images appears to have been initially investigated
in~\cite{zhu2019dlg}, which proposed the Deep Leakage from Gradients
(DLG) algorithm.  An image is input to a neural net and the output is
a label specifying an object detected in the image.  In DLG a
synthetic input is applied to the neural net and the gradients of the
model parameters are calculated.  These gradients are then compared to
the observed model gradients sent to the FL server and gradient
descent is used to update the synthetic input so as to minimise the
difference between its model gradients and the observed model
gradients.

This work was subsequently extended
by~\cite{zhao2020idlg,geiping_inverting_2020-1,wang2020sapag,zhu2020rgap,yin2021see,jin2021catastrophic}
to improve the stability and performance of the original DLG
algorithm, as well as the fidelity of the images it
generates. In~\cite{yin2021see,jin2021catastrophic}, changes to the
optimisation terms allowed for successful data reconstruction at batch
sizes of up to 48 and 100 respectively. Analytical techniques of data
extraction~\cite{boenisch2021curious,pan2020theory} benefit from not
being as costly to compute as compared to optimization based
methods. Additionally, these analytical attacks extract the exact
ground truth data, as compared to DLG and others who often settle to
image reconstructions that include artefacts.

\subsection{Text Data Reconstruction}
Most work on reconstruction attacks has focused on images and there
is relatively little work on text reconstruction.  A single text
reconstruction example is presented in~\cite{zhu2019dlg}, with no
performance results.  Probably the closest work to the present paper
is~\cite{deng2021tag} which applies a variant of DLG to text
reconstruction from gradients of transformer-based models (variants of
BERT~\cite{vaswani2017attention}).  As already noted, DLG tends to
perform poorly with text data and the word recovery rate
achieved in~\cite{deng2021tag} is generally no more than 50\%.

In our attack context, DLG can recover words but at much smaller
scales than we have demonstrated, and takes longer to find these
words. There is also no guarantee that DLG can recover words and place
them in the correct order. In \cite{zhu2019dlg}, it is noted that the
algorithm requires multiple restarts before successful
reconstruction. Additionally, DLG operates by matching the single
gradient of a batch of training data, therefore it only works in the
FedSGD setting, where $E = 1$. We show the results of DLG in Listing
\ref{lst:sampledlg} on gradients of $B = 1, \mbox{ and } 2$ 4 word
sentences. In the first example, it took DLG 1000 iterations to
produce ``\texttt{<S> how are venue}'', and 1500 iterations to produce
``\texttt{<S> how are sure cow}, \texttt{<S> haha where are Tell
  van}''. These reconstructions include some of the original words,
but recovery is not as precise as our attack, and takes orders of
magnitude longer to carry out.

\begin{lstlisting}[frame=single,caption=Original and reconstructed sentences by DLG,label=lst:sampledlg]
  <S> how are you
  <S> how are venue

  <S> how are you doing 
  <S> how are sure cow  
  <S> where are you going
  <S> haha where are Tell van
\end{lstlisting}

Work has also been carried out on membership attacks against text data
models such as GPT2 i.e. given a trained model the attack seeks to
infer one or more training data points.  See for
example~\cite{carlini2019secretsharer,carlini2021extracting}.  But, as
already noted, such attacks are not the focus of the present paper.

\subsection{Proposed Defences}
Several defences have been proposed to prevent data leakage in FL. 

Abadi et al. \cite{abadi2016deep} proposed Differentially Private
Stochastic Gradient Descent (DP-SGD), which clips stochastic gradient
decent updates and adds Gaussian noise at each iteration.  This aims
to defend against membership attacks against neural networks, rather
than the reconstruction attacks that we consider here.
In~\cite{brendan2018learning} it was appled to train a next word
prediction RNN motivated by mobile keyboard applications, again with a
focus on membership attacks.  Recently, the same team at Google
proposed DP-FTRL~\cite{kairouz2021practical} which avoids the sampling
step in DP-SGD.


Secure Aggregation is a multi-party protocol proposed in 2016
by~\cite{bonawitz2016practical} as a defence against data leakage from
the data uploaded by clients to an FL server. In this setting the
central server only has access to the sum of, and not individual
updates.  However, this approach still requires clients to trust that
the PKI infrastructure is honest since dishonest PKI infrastructure
allows the server to perform a sybil attack (see Section 6.2
in~\cite{bonawitz2016practical}) to reveal the data sent by an
individual client.  When both the FL server and the PKI infrastructure
are operated by Google then Secure Aggregation requires users to trust
Google servers to be honest, and so offers from an attack capability
point of view offers no security benefit. Recent work by Pasquini et
al. \cite{pasquini2021eluding} has also shown that by distributing
different models to each client a dishonest server can recover
individual model updates.  As a mitigation they propose adding local
noise to client updates to obtain a form of local differential
privacy.  We note that despite the early deployment of FL in
production systems such as GBoard, to the best our knowledge, there
does not exist a real-world deployment of secure aggregation. This is
also true for homomorphically encrypted FL, and FL using Trusted
Execution Environments (TEEs).

\section{Performance Of Our Attacks Against Federated Learning with Local DP}
Typically, when differential privacy is used with FL noise is added by the server to the aggregate update from multiple clients i.e. no noise is added to the update
before leaving a device.   This corresponds to the situation considered in Section \ref{sec:results_vanilla} .   In this section we now evaluate how local differential privacy, that is, noise added either during local training (DPSGD) or to the final
model parameters $\theta_1$, before its transmission to the
coordinating FL server, affect the performance of both our word
recovery and sentence reconstruction attacks.
\begin{algorithm}
  \caption{Local DPSGD}\label{alg:localdpsgd}
  \KwProc{clientUpdateDPSGD}{
    $\theta_1 \gets \theta_0$\;
    $\mathcal{B} \gets (\mbox{split dataset into batches of size } B)$\;
    $\mbox{\textbf{for} local epoch \textit{i} from 1 to $E$ \textbf{do}}$\;
    $\ \ \mbox{\textbf{for} batch \textit{b} $\in \mathcal{B}$ \textbf{do}}$\;
    $\ \ \ \ \theta_1 \gets \theta_1 - \eta\nabla \ell(\theta_1;b) + \eta\mathcal{N}(0, \sigma)$\;
    \Return $\theta_1$\;
  }
\end{algorithm}

Algorithm \ref{alg:localdpsgd} outlines the procedure for DPSGD-like
local training, where Gaussian noise of mean $0$ and standard
deviation $\sigma$ is added along with each gradient update. Algorithm
\ref{alg:singlenoise} details the typical FL client update procedure
but adds Gaussian noise to the final model $\theta_1$ before it is
returned to the server. In our experiments, everything else as
described in Section \ref{sec:results_vanilla} is kept the same.

\begin{algorithm}
  \caption{Local Single Noise Addition}\label{alg:singlenoise}
  \KwProc{clientUpdateSingleNoise}{
    $\theta_1 \gets \theta_0$\;
    $\mathcal{B} \gets (\mbox{split dataset into batches of size } B)$\;
    $\mbox{\textbf{for} local epoch \textit{i} from 1 to $E$ \textbf{do}}$\;
    $\ \ \mbox{\textbf{for} batch \textit{b} $\in \mathcal{B}$ \textbf{do}}$\;
    $\ \ \ \ \theta_1 \gets \theta_1 - \eta\nabla \ell(\theta_1;b)$\;
    \Return $\theta_1 + \mathcal{N}(0, \sigma)$\;
  }
\end{algorithm}

\begin{figure*}[htb]
  \centering

  \begin{subfigure}[b]{0.315\textwidth}
    \centering
    \includegraphics[width=\textwidth]{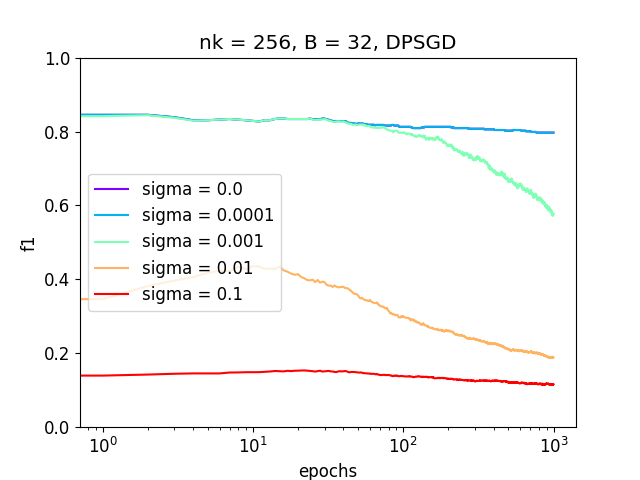}
    \caption{}
    \label{fig:nocutoff}
  \end{subfigure}
    \begin{subfigure}[b]{0.315\textwidth}
    \centering
    \includegraphics[width=\textwidth]{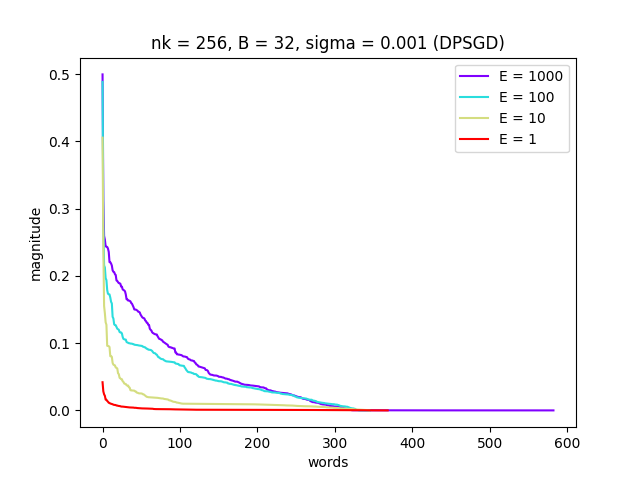}
    \caption{}
    \label{fig:mags1}
  \end{subfigure}
  \begin{subfigure}[b]{0.315\textwidth}
    \centering
    \includegraphics[width=\textwidth]{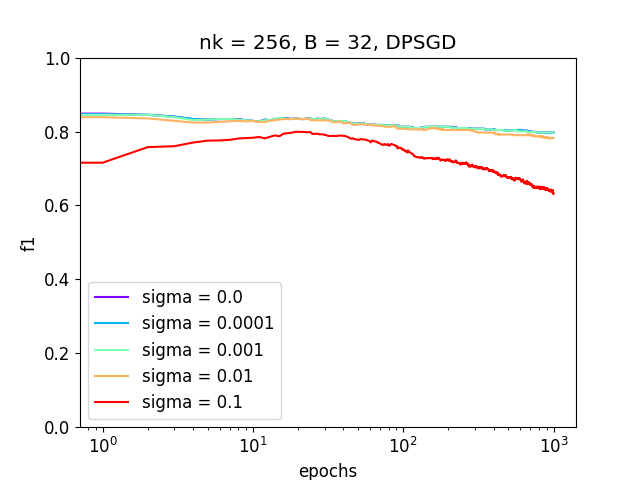}
    \caption{}
    \label{fig:cutoff}
  \end{subfigure}
  \caption{\small Word recovery behaviour when Gaussian noise is all to local FL updates: (a) vanilla word recovery performance, (b) disparity of
    magnitudes between those words that were present in the dataset
    and those 'noisily' flipped negative, (c) word recovery performance when filtering is used.}
  \label{fig:removing-noisy}
\end{figure*}
\begin{figure*}
  \centering
  \begin{subfigure}[b]{0.315\textwidth}
    \centering
    \includegraphics[width=\textwidth]{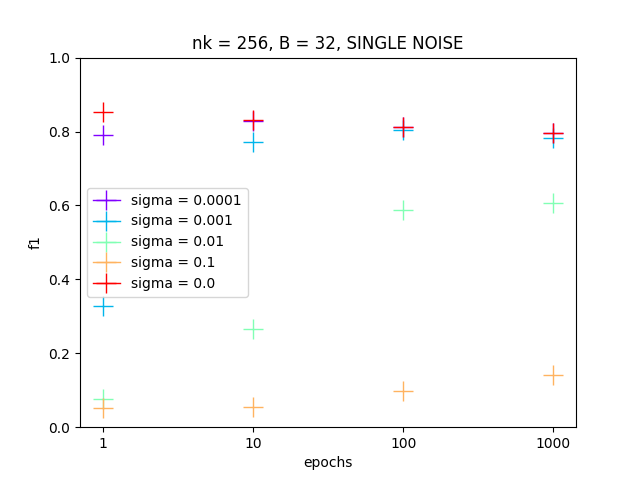}
    \caption{}
    \label{fig:fedavgsinglenoise}
  \end{subfigure}
  \begin{subfigure}[b]{0.315\textwidth}
    \centering
    \includegraphics[width=\textwidth]{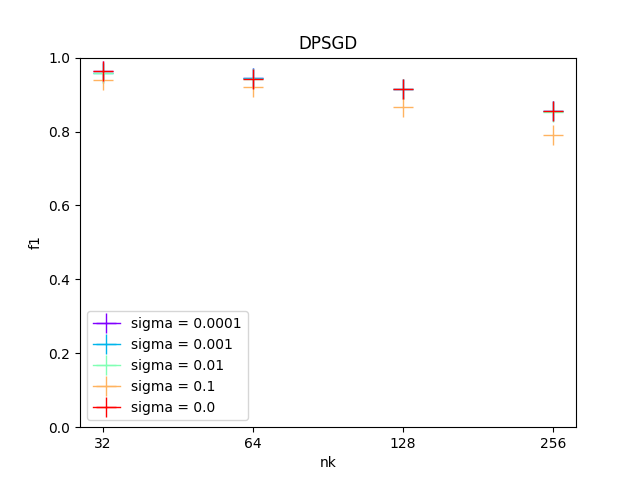}
    \caption{}
    \label{fig:fedsgddpsgd}
  \end{subfigure}
  \begin{subfigure}[b]{0.315\textwidth}
    \centering
    \includegraphics[width=\textwidth]{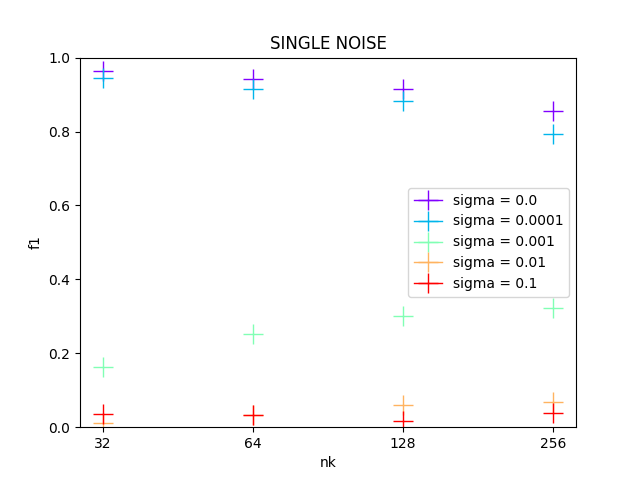}
    \caption{}
    \label{fig:fedsgdsinglenoise}
  \end{subfigure}
  \caption{\small Word recovery results for two local DP
    methods. Figure \ref{fig:fedavgsinglenoise} shows how added noise
    to the final model parameters affects the attack for different
    training times. With DPSGD-like training, noise levels of up to
    $\sigma = 0.1$ are manageable with our magnitude threshold trick,
    resulting in F1 scores close to those had we not added any noise
    at all. With FedSGD updates then for noise of up to $\sigma = 0.0001$
    added to the final parameters, we can still recover words to a
    high degree of precision and recall. }
  \label{fig:noiseresults}
\end{figure*}

\subsection{Word Recovery Performance}

Figure \ref{fig:nocutoff} shows the performance of our word recovery
attack against DPSGD-like local training for different levels of
$\sigma$. For noise levels of $\sigma = 0.001$ or greater, it can be
seen that the F1 score drops significantly.  What is happening is that
the added noise introduces more negative values in the difference of
the final layer parameters and so results in our attack extracting
more words than actually occurred in the dataset, destroying its
precision. However, one can eliminate most of these ``noisily'' added
words by simple inspection. Figure \ref{fig:mags1} graphs the sorted
magnitudes of the negative values in the difference between the final
layer parameters of $\theta_1$ and $\theta_0$, after DPSGD-like
training with $B = 32, n_k = 256, \mbox{ and } \sigma = 0.001$. We can
see that the more epochs the model is trained for, the more words are
extracted via our attack. Of the around 600 words extracted after 1000
epochs of training, only about 300 of them were actually present in
the dataset. On this graph, those words with higher magnitudes
correspond to the ground truth words. It can be seen that we therefore
can simply cutoff any words extracted beyond a specified magnitude
threshold.  This drastically improves the performance of the attack,
see Figure \ref{fig:cutoff}.  It can be seen that even for
$\sigma = 0.1$, we now get word recovery results close to those
obtained when we added no noise at all.

Figure \ref{fig:fedavgsinglenoise} shows the performance of our word
recovery attack when noise is added to the local model parameters
$\theta_1$ in the FedAveraging setting, with
$n_k = 256, \mbox{ and } B = 32$. When $\sigma \ge 0.01$, it can be
seen that the performance drops drastically\footnote{Note that in
  DPSGD the added noise is multiplied by the learning rate $\eta$, and
  so this factor needs to be taken into account when comparing the
  $\sigma$ values used in DPSGD above and with single noise addition.
  This means added noise with standard deviation $\sigma$ for DPSGD
  corresponds roughly to a standard deviation of
  $\eta \sqrt{EB}\sigma$ with single noise addition.  For
  $\eta=0.001$, $E=1000$, $B=32$, $\sigma=0.1$ the corresponding
  single noise addition standard deviation is 0.018.}, even when we
use the magnitude threshold trick described previously.  With FedSGD
(Figures \ref{fig:fedsgddpsgd} and \ref{fig:fedsgdsinglenoise}), we
see that with DPSGD-like training, these levels of noise are
manageable, but for single noise addition, there are only so many
words that are recoverable before being lost in the comparatively
large amounts of noise added.

For comparison, in the FL literature on differential privacy, the
addition of Gaussian noise with standard deviation no more than around
0.001 (and often much less) is typically considered, and is only added
after the update has been transmitted to the coordinating FL server.

\subsection{Sentence Reconstruction Performance}
\begin{figure*}[htb]
  \centering
  \begin{subfigure}[b]{0.45\textwidth}
    \centering
    \includegraphics[width=\textwidth]{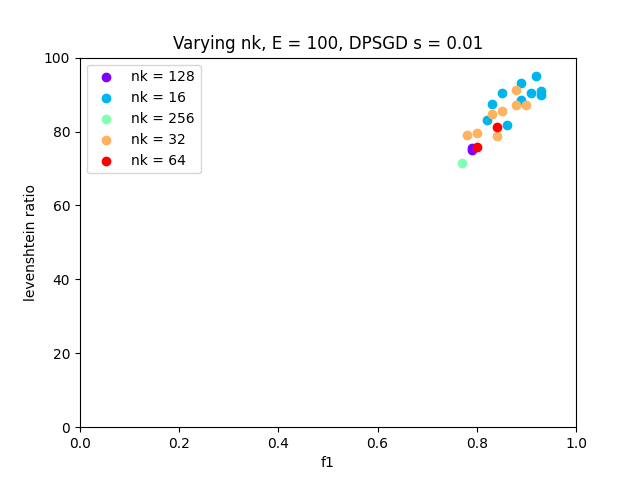}
    \caption{}
    \label{fig:levendpsgd0_01}
  \end{subfigure}
  \hfill
  \begin{subfigure}[b]{0.45\textwidth}
    \centering
    \includegraphics[width=\textwidth]{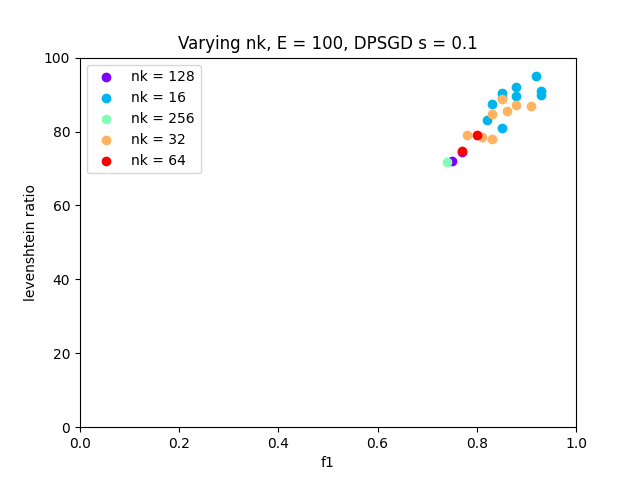}
    \caption{}
    \label{fig:levendpsgd0_1}
  \end{subfigure}
  \vskip\baselineskip
  \begin{subfigure}[b]{0.45\textwidth}
    \centering
    \includegraphics[width=\textwidth]{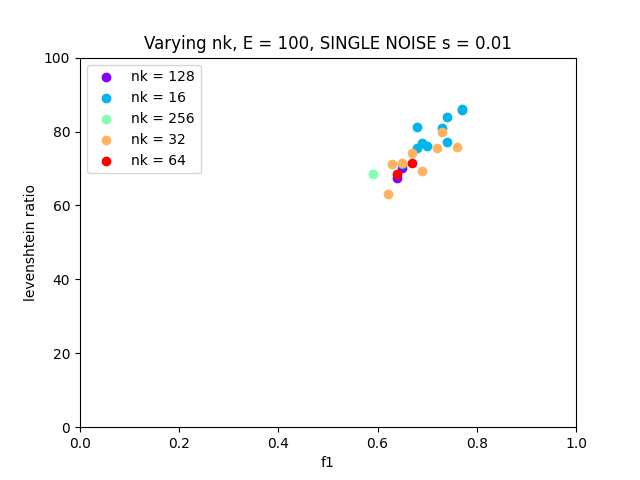}
    \caption{}
    \label{fig:sn0_01}
  \end{subfigure}
  \hfill
  \begin{subfigure}[b]{0.45\textwidth}
    \centering
    \includegraphics[width=\textwidth]{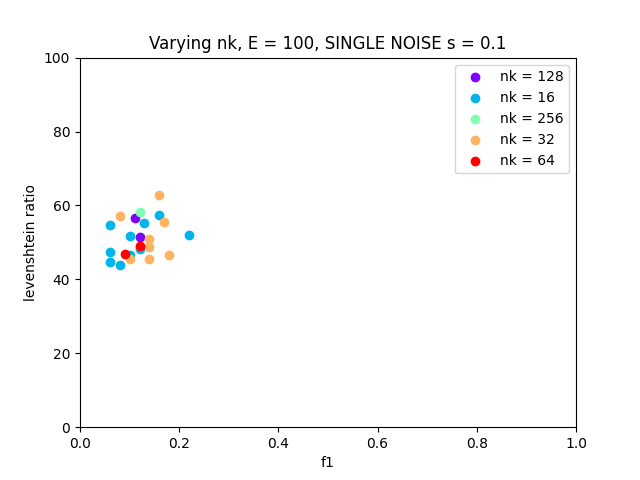}
    \caption{}
    \label{fig:sn0_1}
  \end{subfigure}
  \caption[]
  {\small Sentence reconstruction performance with local DP for
    FedAveraging. The top two Figures (\ref{fig:levendpsgd0_01} and
    \ref{fig:levendpsgd0_1}) show the reconstruction performance for
    different datasets colour coded by their size for DPSGD-like
    training, with $E = 100, \ B = 32$. Here we see no real effect in
    our results compared to the noise-free case. The bottom two figures show the effect of single
    noise addition on sentence reconstruction for the same setting. }
  \label{fig:sr_localdp}
\end{figure*}

Figure \ref{fig:sr_localdp} shows the measured sentence reconstruction performance
with both DPSGD-like training (Figures \ref{fig:levendpsgd0_01} and
\ref{fig:levendpsgd0_1}) and when noise is added to the final
parameters of the model (Figures \ref{fig:sn0_01} and
\ref{fig:sn0_1}). By removing the noisily added words and running our
sentence reconstruction attack, we get results close to those had we
not added any noise for up to $\sigma = 0.1$. For the single noise
addition method, as these levels of noise are not calibrated,
$\sigma = 0.1$ is enough to destroy the quality of reconstructions,
however these levels of noise also destroy any model utility.


\section{Additional Material}
The code for all of the attacks here, the LSTM model and the datasets
used are all publicly available on github
\href{https://github.com/namilus/nwp-fedlearning}{\texttt{here}}.

\section{Summary And Conclusions}
In this paper we introduce two local reconstruction attacks against
federated learning when used to train natural language text models.
We find that previously proposed attacks (DLG and its variants)
targeting image data are ineffective for text data and so new methods
of attack tailored to text data are necessary.  Our attacks are simple
to carry out, efficient, and highly effective.  We illustrate their
effectiveness against the next word prediction model used in Google's
GBoard app, a widely used mobile keyboard app (with $>5$ Billion
downloads) that has been an early adopter of federated learning for
production use.  We demonstrate that the words a user types on their
mobile handset, e.g. when sending text messages, can be recovered with
high accuracy under a wide range of conditions and that
counter-measures such a use of mini-batches and adding local noise are
ineffective.  We also show that the word order (and so the actual
sentences typed) can be reconstructed with high fidelity.  This raises
obvious privacy concerns, particularly since GBoard is in production
use.

Secure multi-party computation methods such as Secure Aggregation and
also methods such as Homomorphic Encryption and Trusted Execution
Environments are potential defences that can improve privacy, but
these can be difficult to implement in practice.  Secure Aggregation
requires users to trust that the server is honest, despite the fact
that FL aims to avoid the need for such trust.  Homomorphic Encryption
implementations that are sufficiently efficient to allow large-scale
production use are currently lacking.

On a more positive note, the privacy situation may not be quite as bad
as it seems given these reconstruction attacks.  Firstly, it is not
the raw sentences typed by a user that are reconstructed in our
attacks but rather the sentences after they have been mapped to tokens
in a text model dictionary.  Words which are not in the dictionary are
mapped to a special \texttt{<UNK>} token.  This means that the
reconstructed text is effectively redacted, with words not in the
dictionary having been masked out.  This suggests that a fruitful
direction for future privacy research on FL for natural language
models may well lie in taking a closer look at the specification of
the dictionary used.  Secondly, we also note that changing from a
word-based text model to a character-based one would likely make our
attacks much harder to perform.


\bibliographystyle{plain}
\bibliography{references}


%

\end{document}